\documentclass{article}

\PassOptionsToPackage{numbers, compress}{natbib}
\usepackage[final]{neurips_2021}
\usepackage{listings}
\usepackage{xcolor}

\definecolor{codegreen}{rgb}{0,0.6,0}
\definecolor{codegray}{rgb}{0.5,0.5,0.5}
\definecolor{codepurple}{rgb}{0.58,0,0.82}
\definecolor{backcolour}{rgb}{0.95,0.95,0.92}

\lstdefinestyle{mystyle}{
    backgroundcolor=\color{white},   
    commentstyle=\color{blue},
    keywordstyle=[1]\color{blue},
    keywordstyle=[2]\color{magenta},
    keywordstyle=[3]\color{blue},
    keywordstyle=[4]\color{codegray},
    numberstyle=\tiny\color{codegray},
    stringstyle=\color{codegreen},
    basicstyle=\ttfamily\scriptsize,
    breakatwhitespace=false,         
    breaklines=true,                 
    captionpos=b,                    
    keepspaces=true,                 
    numbers=left,                    
    numbersep=5pt,                  
    showspaces=false,                
    showstringspaces=false,
    showtabs=false,                  
    tabsize=2,
}
\lstdefinelanguage{mlir}{
    alsoletter={\%,\#,!,.,_},
    morekeywords={\%},
    keywords=[1]{func, return, attributes},
    keywords=[2]{partir.tile, partir.slice, partir.yield, partir.atomic, partir.spmd, !partir.range},
    keywords=[4]{tensor, f32},
    showstringspaces=false,
	breaklines=true,
    breakatwhitespace=true,
    morestring=[b]",
    morecomment=[l]{//},
}
\usepackage[utf8]{inputenc} 
\usepackage[T1]{fontenc}    
\usepackage{hyperref}       
\usepackage{url}            
\usepackage{booktabs}       
\usepackage{amsfonts}       
\usepackage{nicefrac}       
\usepackage{microtype}      
\usepackage{xcolor}         
\usepackage[frozencache,cachedir=.]{minted}

\setminted{fontsize=\small}
\usepackage{subcaption}
\usepackage{graphicx}
\usepackage{wrapfig}

\title{Automap: Towards Ergonomic Automated Parallelism for ML Models}

\author{%
Michael Schaarschmidt~\thanks{Correspondence: mschaarschmidt@deepmind.com.} \qquad
  Dominik Grewe \qquad
  Dimitrios Vytiniotis \qquad
  Tamara Norman \AND
  James Molloy \qquad
  Jonathan Godwin \qquad
  Norman A. Rink \qquad
  Vinod Nair \qquad
  Dan Belov \\
  DeepMind \\
  \And 
 Adam Paszke \\
  Google Research \\
  \And
  Georg Stefan Schmid \\
  EPFL \\
}

\begin{document}

\maketitle

\begin{abstract}
The rapid rise in demand for training large neural network architectures has brought into focus the need for partitioning strategies, for example by using data, model, or pipeline parallelism. Implementing these methods is increasingly supported through program primitives, but identifying efficient partitioning strategies requires expensive experimentation and expertise. We present the prototype of an automated partitioner that seamlessly integrates into existing compilers and existing user workflows. Our partitioner enables SPMD-style parallelism that encompasses data parallelism and parameter/activation sharding. Through a combination of inductive tactics and search in a platform-independent partitioning IR, automap can recover expert partitioning strategies such as Megatron sharding for transformer layers.
\end{abstract}

\section{Introduction}
Driven by recent progress in the design and evaluation of large language models \cite{gpt3_2020, gshard, scaling_laws_2020, codex2021}, training techniques for large deep neural networks have become critically important for progress in machine learning systems. These networks are trained by combining multiple parallelism strategies and executing them across many accelerator devices. Identifying an effective combination of approaches such as data, model \cite{megatron2019} or pipeline parallelism \cite{gpipe, pipedream, Narayanan2021MemoryEfficientPD, pipemare21} depends on the specific model architecture, accelerator characteristics, and distributed device topology. Selecting from a growing array of techniques such as micro-batching, rematerialisation, or parameter offloading \cite{DBLP:journals/corr/abs-2101-06840} is further complicated by expensive experimentation, with large models requiring up to thousands of accelerators. Beyond training larger models, organisations managing diverse accelerator fleets can improve hardware utilisation by partitioning models to fit onto older accelerators with less memory. 

Tensor programming frameworks like TensorFlow \cite{tensorflow2015-whitepaper}, PyTorch \cite{pytorch2019} and JAX \cite{jax2018github} increasingly provide end-user primitives or add-in libraries to help define the parallelisation strategy. JAX exposes several function transformations to control parallelism such as \textit{pmap} (typically, but not exclusively, used for batch parallelism), \textit{pjit} or \textit{xmap} for fine-grained control of model parallelism and interfacing tools such as the XLA SPMD partitioner \cite{gshard, gspmd2021}. While these are powerful tools which enable experts to compose advanced parallelism strategies, they still require specialised skill sets to achieve the desired hardware utilisation. For instance, users must come up with expert sharding specifications for parameters and possibly
intermediate values of a model to productively use the \textit{pjit} API, which in
turn drives the XLA SPMD partitioning infrastructure.

\textbf{Motivation and challenges.} Though automated partitioning and distribution of ML workloads has been explored before in the research community~\cite{deviceplacement2017, optgraphcomp, mirhoseini2018a, flexflow2019, roc2020, tofu2019, mirhoseini2018a}, we present here the unique challenges motivating our work:

\begin{itemize}
\item We need integration into existing backend compilers/runtimes widely used in production and already targeting different accelerator platforms, most notably XLA~\cite{xla}. We explicitly want to avoid having to re-implement kernels for a specific architecture to be partition-aware~\cite{flexflow2019}.
\item We need integration into existing user workflows, in our case {\em arbitrary} computations specified in JAX without {\em any} user rewriting. Our partitioner may have to deal with XLA programs consisting of hundreds of thousands of operators, with hundreds or thousands of parameters.
We cannot use a predefined library of partitioning-aware layers (e.g. Keras layers, as in~\cite{flexflow2019}) because we wish to allow researchers (our prevalent group of users) to freely innovate with arbitrary tensor computations and experimental layers.
\item We aim for a fast solution that allows an effective research development cycle, i.e. a solution comparable to the overhead to schedule an experiment, perhaps minutes but not hours.
\end{itemize}

\textbf{Key ideas we offer for discussion.}
In this paper, we present our design prototype and preliminary results of a data-driven automatic partitioner, \textit{automap}. The challenging setting we face has implications (a) on the design of our partitioning infrastructure to make it compatible with existing compilers and user workflows, and (b) on the design of our automation. 

To address problem (a) we offer a novel IR (``dialect'' in MLIR terms~\cite{mlir2020}) that is {\em layered} on top of another tensor dialect (XLA HLO in our prototype, but this could be adapted to any dialect) and allows us to express distribution decisions of computations from the base dialect. The distribution decisions are expressed as {\em rewrite
rules} controlled by an agent, providing maximal flexibility to incorporate new rewrites (Section~\ref{partir}).

Problem (b) is particularly challenging: The most important implication of our setup is the need to manage large, unstructured programs. Contrary to coarse-grained layer-based approaches, it would be impractical to explore decisions\footnote{A decision may be to determine whether a value (e.g. an argument or an intermediate) should be replicated or sharded, and if the latter, upon which dimension and along which devices (mesh axis).} for all operations in a program via search. On the other hand, ML models can be trained  to
make parallel decisions for all operations \cite{optgraphcomp} at this scale, but such models typically rely on fine-tuning for new types of programs, and this makes the approach harder to integrate with a researcher's workflow. Moreover, evaluating the ``goodness'' of a partitioning solution, e.g., the reduction in peak working memory, requires at least a static analysis (e.g. a liveness analysis) over the {\em result of lowering and optimising} a large (50-100k ops) program to an accelerator-local program. This reinforces the need for solutions that get to good performance within few trials and rewriting steps. Hence, we started our investigations by exploring (i) hybrid search-learning techniques, and (ii) incremental partitioning decisions and the use of compiler rewriting passes that propagate these decisions across the program where possible. A key idea we use is to {\em imitate the behaviour of expert users partitioning their models} (e.g. users of GSPMD~\cite{gspmd2021}) to design inductive biases that reduce the number of decisions taken by an agent. Early results show how this approach can recover expert partitioning such as the Megatron transformer layer sharding~\cite{megatron2019}. We identify important challenges for scaling search up to many layers and improving its robustness to the details of the model architecture.

\begin{figure}
\centering
 \includegraphics[scale=0.4]{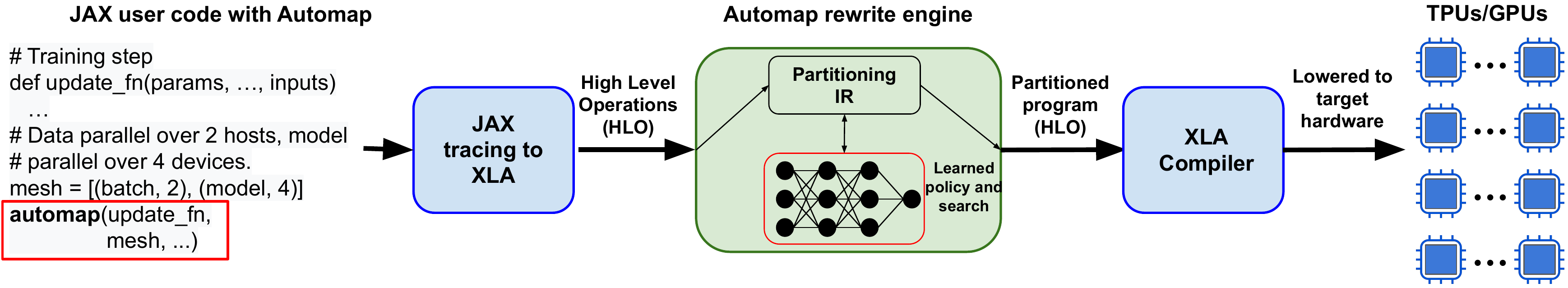}
\caption{End-to-end workflow overview. \label{fig:overview}}
\vspace{-0.5cm}
\end{figure}

\textbf{Further related work.} There is further related work, not discussed in detail in this paper -- for instance DistIR \cite{distir2021} is also an MLIR-based representation
for distributed tensor programs focusing on MPMD~(for arguments for/against MPMD see~\cite{gspmd2021}). ROC~\cite{roc2020} is an extension of~\cite{flexflow2019} specifically for GraphNets using a learned cost model. Tofu~\cite{tofu2019} is a hierarchical graph partitioner based on dynamic programming. Other plausible techniques for partitioning such as constraint-based encoding of cost models, often combined with equality saturation~\cite{equality_saturation21}, or BRKGA~\cite{aditya2020}-based options are all insufficient -- at least out of the box -- for our purposes due to the difficulties in obtaining cost models. Another promising avenue is to design {\em progressively accurate} sequences of cost models that do not all require a fully materialised program, as in TVM~\cite{tvm2018}.

\section{System design}
Our rewrite engine is implemented in MLIR \cite{mlir2020} with an XLA backend, and a Python API in JAX. Users interact with our system by designing normal JAX models and then pass their main update function to our partitioner. The JAX functions can already include user-managed parallelism (e.g. batch parallelism), and our system will further partition them along additional mesh axes. JAX functions are converted to XLA computations, and then lowered to our rewriting dialect PartIR \cite{partir2020} in MLIR (overview in Figure~\ref{fig:overview}). Rewrites and inductive tactics are exposed to the partitioner via a compiler environment (\S\ref{partitioner}). Rewrite sequences executed by the partitioner are lowered to an SPMD variant of our partitioning IR, and evaluated through compiler-internal cost models (estimating peak memory, runtime, and communication). PartIR is platform-independent but we have implemented an XLA backend to seamlessly support CPU, TPU \cite{DBLP:journals/corr/JouppiYPPABBBBB17}, and GPU worfklows.

\begin{figure}[h]\scriptsize
\begin{tabular}{c}
\begin{minipage}{\textwidth}
\begin{lstlisting}[language=mlir]
func @main(%arg0: tensor<8x16xf32>, %arg1: tensor<16x64xf32>, %arg2: tensor<64xf32>) 
  -> tensor<8x64xf32> {
    %0 = mhlo.dot %arg0, %arg1 : tensor<8x64xf32>
    %1 = mhlo.broadcast_in_dim %arg2 {broadcast_dims = 1} : tensor<8x64xf32>
    %2 = mhlo.add %0, %1 : tensor<8x64xf32>
    return %2 : tensor<8x64xf32>
}
\end{lstlisting}
\end{minipage}\\\hline\\
\begin{minipage}{\textwidth}
\begin{lstlisting}[language=mlir]
func @main(%arg0: tensor<8x16xf32>, %arg1: tensor<16x64xf32>, %arg2: tensor<64xf32>) 
  -> tensor<8x64xf32> attributes {mesh_shape = #partir.mesh<"shard"=2>} {
    %0 = partir.tile 1 "shard" (%rshard : !partir.range<2>) {
      %4 = partir.slice 1 %arg1[%rshard] : tensor<16x32xf32>
      partir.yield %4 : tensor<16x32xf32>
    }
    %1 = mhlo.dot %arg0, %0  : tensor<8x64xf32>
    %2 = mhlo.broadcast_in_dim %arg2 {broadcast_dims = 1} : tensor<8x64xf32>
    %3 = mhlo.add %1, %2 : tensor<8x64xf32>
    return %3 : tensor<8x64xf32>
}
\end{lstlisting}
\end{minipage}\\ \hline\\
\begin{minipage}{\textwidth}
\begin{lstlisting}[language=mlir]
func @main(%arg0: tensor<8x16xf32>, %arg1: tensor<16x64xf32>, %arg2: tensor<64xf32>) 
  -> tensor<8x64xf32> attributes  {mesh_shape = #partir.mesh<"shard"=2>} {
    %0 = partir.atomic "shard" {
      partir.yield %arg0 : tensor<8x16xf32>
    }
    %1 = partir.tile 1 "shard" (%rshard : !partir.range<2>) {
      %2 = partir.slice 1 %arg1[%rshard] : tensor<16x32xf32>
      %3 = mhlo.dot %0, %2 : tensor<8x32xf32>
      %4 = partir.slice 0 %arg2[%rshard] : tensor<32xf32>
      %5 = mhlo.broadcast_in_dim %4 {broadcast_dims = 1} : tensor<8x32xf32>
      %6 = mhlo.add %3, %5 : tensor<8x32xf32>
      partir.yield %6 : tensor<8x32xf32>
    }
    return %1 : tensor<8x64xf32>
}
\end{lstlisting}
\end{minipage}
\end{tabular}
\caption{Top: A small MLIR MHLO program representing a single linear layer in slightly simplified
notation. Middle: the same program where {\tt \%arg1} has been expressed as a tiling loop on dimension 1. Bottom: the final PartIR program after propagation. Note that looping on dimension 1 of {\tt \%arg1} (of size 64) means that we can also partition the dot product along dimension 1, and essentially pull the whole computation inside the tiling 
loop by operating on slices of {\tt \%arg2}. In the final program {\tt \%arg0} automatically got wrapped inside an ``atomic'' region to signify that it will remain replicated.\label{partir-mlir}}
\end{figure}

\subsection{Partitioning IR}\label{partir}
To expose partitioning decisions, we represent tensor programs in PartIR which is an MLIR ``dialect'' layered
on top of MHLO, an MLIR encoding of the XLA Higher Level Operations (HLO). At the core of our IR, which operates on statically shaped multi-dimensional arrays, are tiling loop-like operators which express parallel computations that compute a {\em distributed} tensor value. Instead of allowing unrestricted parallel loops, we force users to declare logical mesh axes with fixed sizes and make sure that every such loop in a program is associated with an axis and same-axes loops never occur nested. This guarantees that our programs can compile as a single SPMD kernel. An example of a $2{\times}4$ mesh is given in the left of Figure~\ref{fig:overview}, requiring a total of 8 devices for execution. 

Rewriting actions include actions to express the distribution of an intermediate variable as well as several flavors of propagation of partitioning information (i) from operands to results; (ii) from results to operands, and (iii) from a subset of operands to the rest. These propagation rules are enabled by a {\em registry} containing a declarative specification of this behaviour for each operator in the underlying tensor dialect. Rewrites always preserve semantics, decoupling search policies from correctness. 

Figure \ref{partir-mlir} illustrates a small MHLO program representing a dense layer, how tiling decisions are expressed in our IR, and the result of propagation. Finally, the tiling loops in our IR lower to a dialect suitable for expressing SPMD computations -- Figure~\ref{partir-spmd} shows the result of lowering the 
final program in Figure~\ref{partir-mlir}. Optimising data transfers and reasoning about cost happens at this level of the stack, before we eventually compile back to accelerator-specific HLO code and feed back into the XLA compiler/runtime. A detailed technical exposition of our IR stack is the subject of a different paper.

\begin{figure}[h]
\begin{lstlisting}[language=mlir]
func @main(%arg0: f32[8,16], %arg1: f32[16,64{"shard"}]>, %arg2: f32[64{"shard"}]>) 
    -> f32[8,64{"shard"}]>  attributes {mesh_shape = #partir.mesh<"shard"=2>} {
    %0 = partir.spmd(%arg1, %arg0, %arg2) ["shard"] (%rshard: !partir.range<2>, 
                                                     %arg3: tensor<16x32xf32>, 
                                                     %arg4: tensor<8x16xf32>, 
                                                     %arg5: tensor<32xf32>) {
      %2 = mhlo.dot %arg4, %arg3 : tensor<8x32xf32>
      %3 = mhlo.broadcast_in_dim %arg5 { broadcast_dims = 1 } : tensor<8x32xf32>
      %4 = mhlo.add %2, %3 : tensor<8x32xf32>
      partir.yield %4 : tensor<8x32xf32>
    }
    return %0
}
\end{lstlisting}

\caption{An SPMD program after partitioning using axis {\em``shard''}. A {\em distributed tensor type} like
{\tt f32[16, 64\{"shard"\}]} means that the value has a global shape of {\tt [16, 64]}, but is 
nevertheless sharded in chunks of shape {\tt [16, 32]} (since here axis {\tt "shard"} is of size 2).~\label{partir-spmd}}
\end{figure}

\subsection{Automated partitioner}\label{partitioner}
The partitioner interacts with a rewriting environment exposing rewriting tactics to distribute/partition values, and tactics involving patterns that apply globally throughout the module. Prior work \cite{optgraphcomp, placeto} has highlighted the cost of auto-regressive rewriting which scales unfavourably with model size. Our approach is guided by minimising the number of rewriting decisions. To improve robustness across a wide range of programs, we propose to combine search and learning based on several observations:
\begin{itemize}
    \item Users often do not need to solve heterogeneous partitioning problems but typically map programs to rectangular device meshes such as TPU slices. Faster automation is enabled by restricting partitioning to using pre-defined mesh axes as the structure of tiling loops and their allowable nesting is fixed ahead of time.
    \item When developing a partitioning strategy, users can often assign some decisions themselves based on knowledge of model and devices, such as selecting a data parallel axis. This allows the partitioner to focus on difficult decisions such as only the model-parallel strategy.
    \item Experts do not approach partitioning by investigating individual operations but consider key structural elements such as parameters, certain activations or inputs, optimiser/network state etc. to formulate a high level strategy. They often then hand-craft annotations for a handful of internal nodes that they deem important to guide tools like the XLA GSPMD partitioner.
\end{itemize}

We leverage all three observations to design our automated partitioner. For a physical set of devices, (e.g. 8), users explicitly specify the logical axes for different forms of parallelism, concretely by providing a set of axes names and sizes (e.g. $\{$(“batch”, 2), (“model”, 4)$\}$. The partitioner only searches over tiling decisions involving axes which it is explicitly instructed to use -- while users remain in control of the others.
\begin{figure}
\begin{minipage}{0.45\textwidth}
\begin{minted}{python}
def update(params, opt_state, batch):
  loss, grads = jax.value_and_grad(
    loss_fn)(params, batch)
  grads = jax.lax.psum(grads, "batch")
  updates, opt_state = opt.update(grads, 
    opt_state)
  params = optax.apply_updates(params, 
    updates)
  return params, opt_state, loss

# Update calls now execute in SPMD
# fashion.
update = jax.pmap(update, 
                  axis_name="batch")
\end{minted}
\caption{Using JAX's {\em pmap} (parallel map) function transform enables the use of multi-host data parallelism over an axis {\em"batch"}.
\label{fig:pmap-example}}
\end{minipage}
\hfill
\begin{minipage}{0.45\textwidth}
\begin{minted}{python}
def update(params, opt_state, batch):
    # Same as in Fig. 2.

# Specify mapping of axes on devices.
device_layout = np.reshape(
    np.array(jax.devices()),  (2, 4))
mesh = Mesh(device_layout ,
            ("batch", "model"))

# Manual data-, automated model parallel.
# Axis "batch" is specified  
# for batch, the third input argument.
update, spec = automap(
  update, mesh, ['model'], (None,
  None, 0) (None, None, 0))(*args)
\end{minted}
\caption{Automap allows combining manual and automated parallelism. Users specify a mesh layout and where manual axes apply.
\label{fig:automap-example}}
\end{minipage}
\vspace{-5mm}
\end{figure}

\subsection{Search and learning}
The partitioner must select a sequence of rewrites and propagation tactics to optimise some
desirable cost function such as the execution cost. Large models frequently reach 50-100k HLO operations, and even the set of interesting nodes, in our experience $\approx 1\%$ of operations, can be impractically large for search. Instead of opting for a fully learned solution which may require fine-tuning on unseen programs, we first experiment with a hybrid approach. A learner narrows down the most relevant subset of program nodes for partitioning, and search selects the final decisions.

\textbf{Search.} We implemented Monte Carlo Tree Search (MCTS) \cite{Browne12asurvey} with upper confidence bound for trees (UCT). Instead of exposing all program operations for assigning mesh axes, we initialise a worklist of all ‘interesting operation nodes’ when traversing the program i.e., the function arguments to the MLIR representation consisting of weights and biases, optimiser state, and model inputs. The action space exposes actions to insert tiling loops that partition each tensor and dimension by each predefined axis. After applying an action, its consequences are propagated {\em conservatively} backward and forward through the program -- for instance if a pointwise operator receives equi-partitioned arguments it may also be executed to produce an equi-partitioned result. Propagation can get stuck in internal nodes for which insufficient information exists (e.g. not enough arguments are partitioned); and these internal nodes with a need for non-trivial decisions resurface back to our worklist. This is a key difference compared to the heuristics-based sharding propagation underlying XLA GSPMD~\cite{gshard}. Another global rewriting decision we expose is a pass that infers the tiling of the rest of the arguments from only some of them. This pass allows us with only a few tiling decisions for some parameters/inputs of a model to induce sharding for other parameters/inputs. If applied at every step, it quickly reduces the number of remaining decisions at the cost of increased wall clock time, and we are experimenting with different mechanics to expose this to an agent.

\textbf{Learning.} Using the worklist still exposes too many nodes for search, but careful analysis of the state-of-the-art sharding of transformers~\cite{megatron2019} shows that in certain situations only a handful need to be selected to fully partition a model, so we apply learning to rank them. Our compiler featurises operation nodes as a concatenation of operation type, operand shapes, and existing partitioned axes. Edges encode program dataflow and MLIR program structure. We then compute a per-node relevance score using a node-embedding where a learned model predicts for each input to the MLIR program a ranking corresponding to the importance of this node to be partitioned, and the top-$k$ ($k =$ 25) most relevant nodes are then passed to MCTS to select the final rewriting sequence. In summary, we combine inductive propagation tactics, search and learning to deliver automated model parallelism with a relatively small number of explicit decisions.

\textbf{End-to-end user example.}  We illustrate how end-users interact with automap in Figure \ref{fig:automap-example} in comparison to using an existing JAX parallel primitive in Figure \ref{fig:pmap-example}. Automap is instrumented using existing JAX tooling for describing positional axes and meshes (based on \textit{xmap}/\textit{pjit}). In addition to a partitioned callable, automap returns a specification of partitioning decisions for inputs and outputs. These specifications can then be used to partition function inputs such as parameters, optimiser state, or network state.

\section{Towards an evaluation}

\textbf{Results.} We investigate the performance of our prototype on a transformer model and compare to a well known reference strategy. We implemented a GPT-3 \cite{gpt3_2020} style 24-layer transformer model which requires $\approx$26 GB of memory at batch size 1 (not fit for a single TPU v3 device at 16 GB RAM), and which has just over 50k operations, and 1150 arguments. We then evaluated our prototype's ability to discover Megatron-style \cite{megatron2019} sharding through search, and when combined with a learner. The search mechanism is guided by multiple cost statistics. First, a peak liveness analysis exposes an approximate memory estimate. This is a conservative estimate, and XLA compilation can further improve required memory through optimisations such as fusion. Second, we minimise the number of bytes communicated through reduction operations. The learned model was trained on a dataset of 20k transformer variants. To generate training data, we selected random model arguments (1000 per model), and exhaustively partitioned all argument dimensions. Our model was trained to imitate the highest scoring strategy. The model was implemented based on an Interaction Network \cite{DBLP:journals/corr/BattagliaPLRK16} using JAX with Haiku, Jraph for GraphNets, and Optax for training \cite{jax2018github, deepmind2020jax, jraph2020github, haiku2020github}.

Megatron is a highly scalable \cite{large_megatron_21} training strategy for transformers \cite{DBLP:journals/corr/VaswaniSPUJGKP17} which exploits intra-layer model parallelism to minimise the number of required all-reduces. We view Megatron as a representative for a widely used expert strategy. Achieving Megatron is measured through gathering statistics on collectives in the partitioned model.

\begin{figure}[h]
\begin{minipage}{0.45\textwidth}
 \includegraphics[scale=0.45]{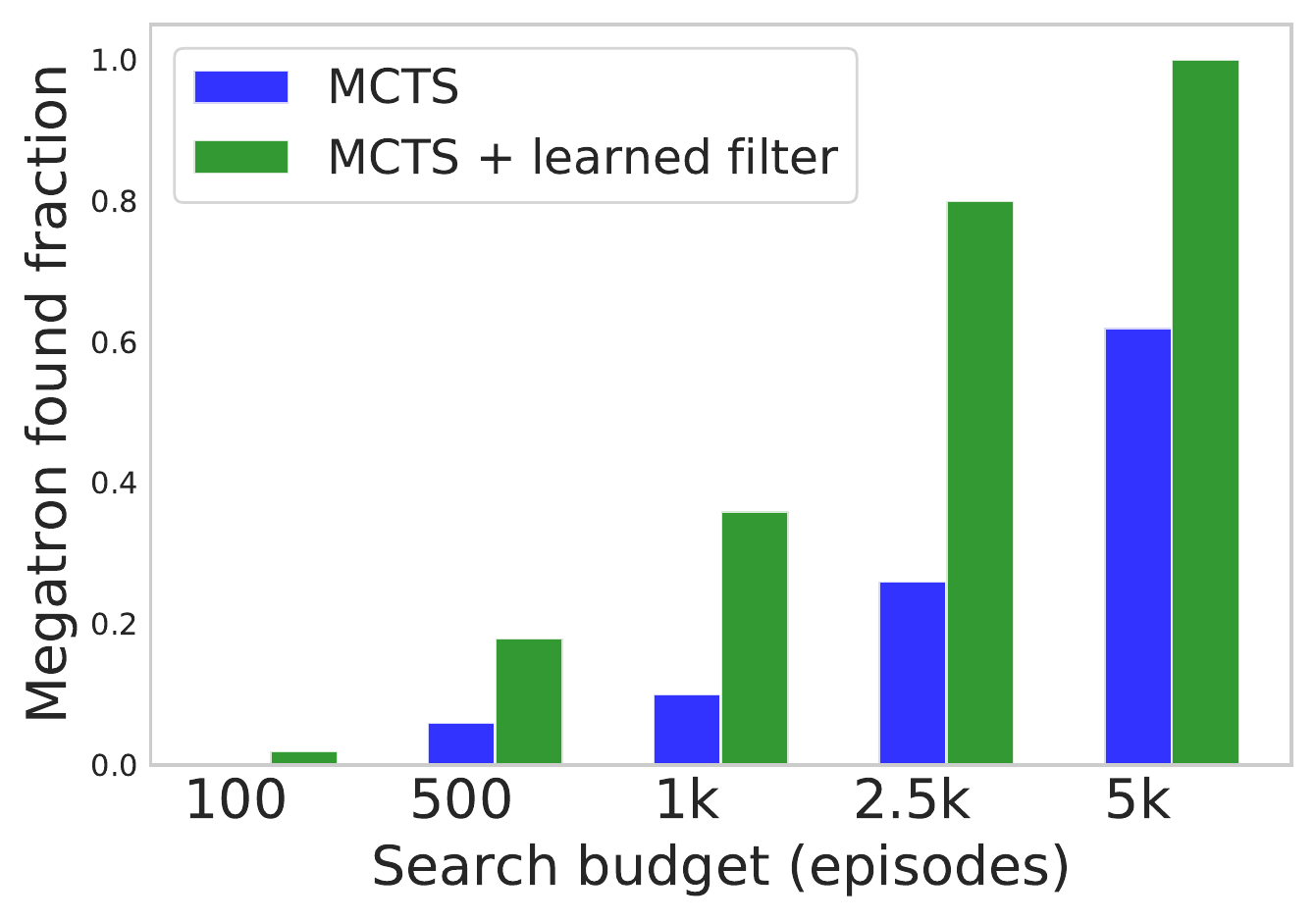}
\caption{Comparing search progress using MCTS only and combined with a learned filter to discover Megatron-style sharding. \label{fig:megatron_fractions}}
\end{minipage}
\hfill
\begin{minipage}{0.45\textwidth}
 \includegraphics[scale=0.45]{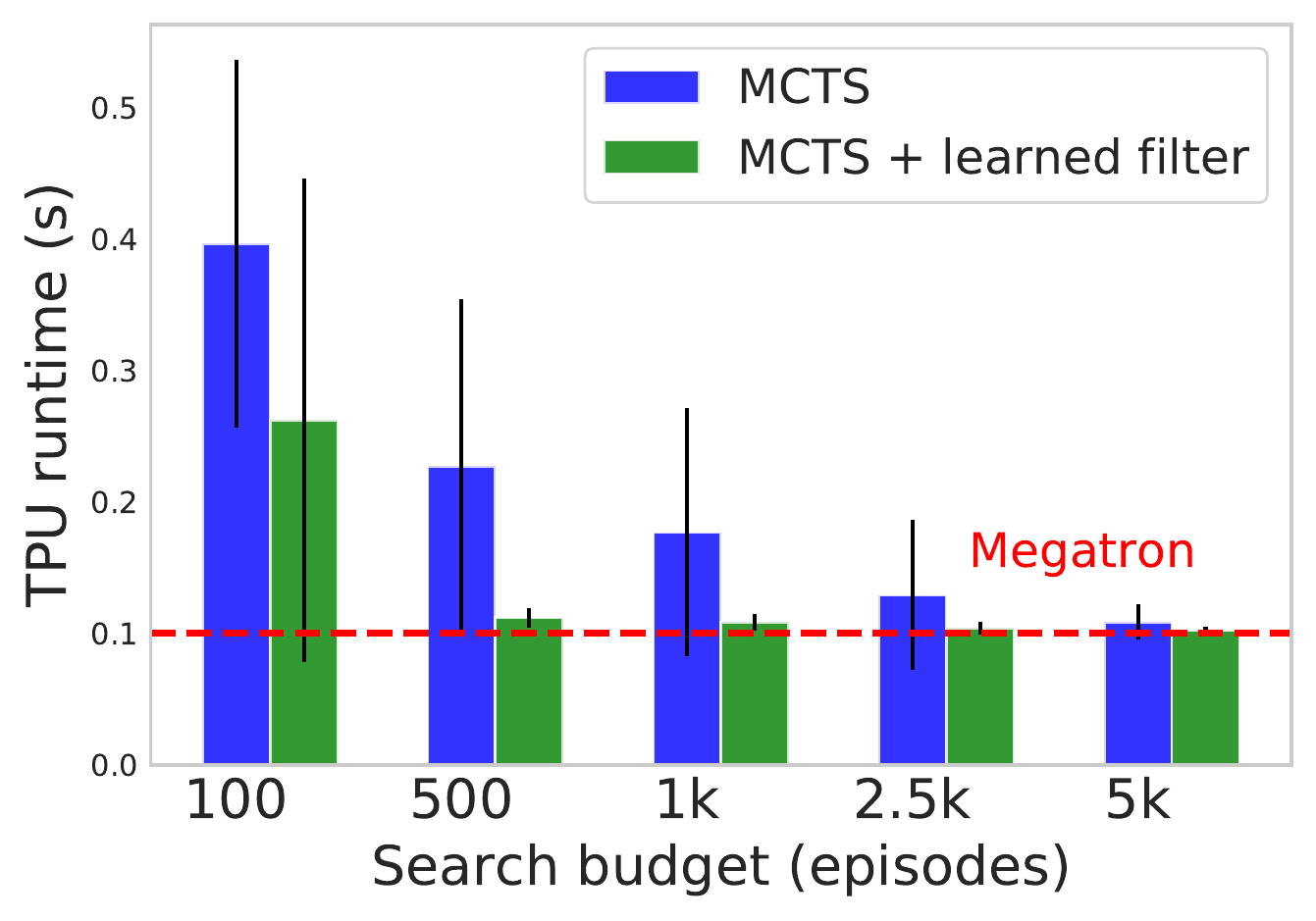}
\caption{TPU v3 runtimes of the solutions found. Near Megatron solutions only incur a small performance penalty. \label{fig:megatron_runtimes}}
\end{minipage}
\end{figure}
In Figure \ref{fig:megatron_fractions} we illustrate the success rate in discovering Megatron (over 50 search attempts) for a number of search budgets. Results show that several thousands of episodes are required to reliably discover expert-level sharding. We then evaluated the search result for each run on TPU v3 (Figure \ref{fig:megatron_runtimes}). A key insight is that our search at shorter budgets frequently discovers solutions near Megatron (i.e. few redundant collectives) which are in practice almost as fast (as highlighted by the runtimes of solutions using the learned filter, which are near Megatron from 500 episodes onwards, i.e. requiring few minutes of search). Solutions typically required 2-20 decisions. While these results are encouraging, more work is needed to support a learned system in interactive compiler workflows to be able to handle a variety of (generally unpredictable) user programs.

\begin{figure}[h]
\begin{minipage}{0.45\textwidth}
 \includegraphics[scale=0.4]{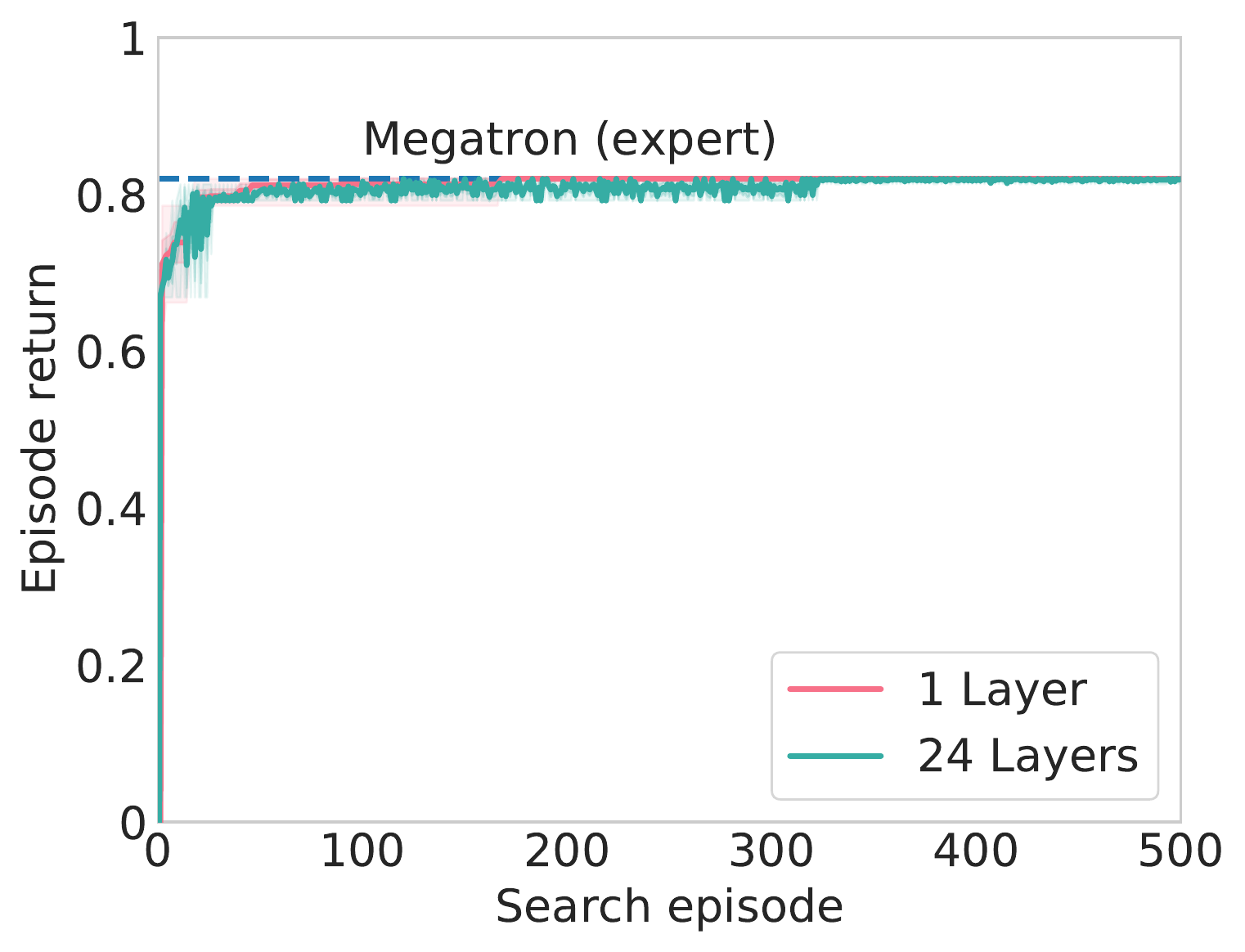}
\caption{Searching Transformer sharding strategies with grouped attention blocks via compiler hints drastically improves results. \label{fig:transformers_grouped}}
\end{minipage}
\hfill
\begin{minipage}{0.45\textwidth}
 \includegraphics[scale=0.4]{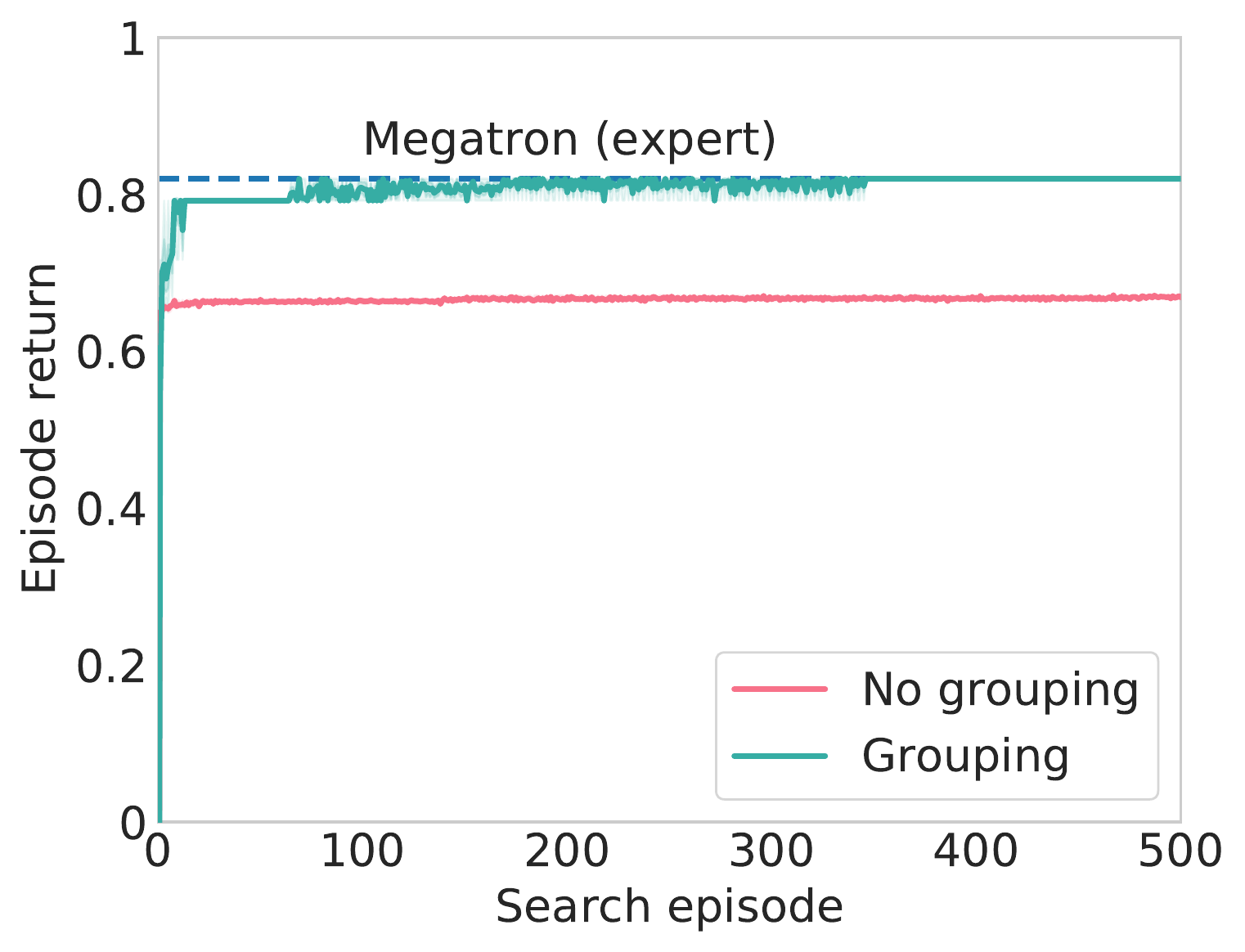}
\caption{Impact of grouping when not relying on propagation of sharding information via shared constants across layers. \label{fig:grouping_comparison}}
\end{minipage}
\end{figure}

\textbf{Scaling with compiler hints.} Our results show that discovering semantically meaningful strategies is possible in principle. However, we found that discovering these strategies for larger models critically relies on propagating sharding information through subtly shared constants and other
computations {\em across layers}. Such sharing is brittle and cannot be relied on for a general solution towards rewriting deeper networks. As machine learning models commonly consist of repeated blocks (such as attention blocks in Transformers, residual blocks in ConvNets, or message passing layers in Graph nets), search techniques scale unfavourably when having to explicitly rewrite each layer. We therefore also implemented the ability to exploit model structure in automap by allowing users to group repeated layers together and exposing only a single set of decisions per group. The mechanism used was that of {\em named scopes} which are commonly used in libraries such as Haiku \cite{deepmind2020jax}. Figures \ref{fig:transformers_grouped} and \ref{fig:grouping_comparison} present the effect of grouping.

Notably, when allowing for compiler hints on layer groups, Megatron can be found reliably in a small number of episodes without requiring to propagate through shared dependencies across layers. Without grouping or shared dependency propagation, Megatron is not found for a 24 layer Transformer. As grouping only requires users to provide the name scope for any relevant group (e.g. "attention-block"), this provides an attractive path for initial real world use cases.

\textbf{Other models.} We were also able to partition other models such as GraphNets where no one-size-fits-all expert strategy exists. Here, the automap prototype in first experiments was able to discover simple manual strategies such as input edge sharding that allow practitioners to begin experimentation with larger graphs and models. 

\textbf{Discussion.} Our results illustrate how effective partitioning strategies can be reached using a rewriting system through the combination of data-driven strategies and inductive tactics, and optionally exploiting high-level model structure. More work is required to understand the right combination of user-provided structure information, search, and learning in order to balance minimal user efforts and time to good solutions. The results presented here were also initially restricted to sharding within the devices of a single host while assuming data parallelism across hosts which simplifies the communication cost model. More advanced cost models will be required to model multi-host communication as well as down-stream changes to models during device-specific lowering (e.g. through fusion). Finally, further work is also needed to support additional automated partitioning strategies such as pipeline parallelism or ZeRO offloading.
\bibliography{biblio}

\begin{thebibliography}{36}
\providecommand{\natexlab}[1]{#1}
\providecommand{\url}[1]{\texttt{#1}}
\expandafter\ifx\csname urlstyle\endcsname\relax
  \providecommand{\doi}[1]{doi: #1}\else
  \providecommand{\doi}{doi: \begingroup \urlstyle{rm}\Url}\fi

\bibitem[xla(2017)]{xla}
{XLA}: Optimizing compiler for machine learning, 2017.
\newblock URL \url{https://www.tensorflow.org/xla}.

\bibitem[Abadi et~al.(2015)Abadi, Agarwal, Barham, Brevdo, Chen, Citro,
  Corrado, Davis, Dean, Devin, Ghemawat, Goodfellow, Harp, Irving, Isard, Jia,
  Jozefowicz, Kaiser, Kudlur, Levenberg, Man\'{e}, Monga, Moore, Murray, Olah,
  Schuster, Shlens, Steiner, Sutskever, Talwar, Tucker, Vanhoucke, Vasudevan,
  Vi\'{e}gas, Vinyals, Warden, Wattenberg, Wicke, Yu, and
  Zheng]{tensorflow2015-whitepaper}
Mart\'{\i}n Abadi, Ashish Agarwal, Paul Barham, Eugene Brevdo, Zhifeng Chen,
  Craig Citro, Greg~S. Corrado, Andy Davis, Jeffrey Dean, Matthieu Devin,
  Sanjay Ghemawat, Ian Goodfellow, Andrew Harp, Geoffrey Irving, Michael Isard,
  Yangqing Jia, Rafal Jozefowicz, Lukasz Kaiser, Manjunath Kudlur, Josh
  Levenberg, Dandelion Man\'{e}, Rajat Monga, Sherry Moore, Derek Murray, Chris
  Olah, Mike Schuster, Jonathon Shlens, Benoit Steiner, Ilya Sutskever, Kunal
  Talwar, Paul Tucker, Vincent Vanhoucke, Vijay Vasudevan, Fernanda Vi\'{e}gas,
  Oriol Vinyals, Pete Warden, Martin Wattenberg, Martin Wicke, Yuan Yu, and
  Xiaoqiang Zheng.
\newblock {TensorFlow}: Large-scale machine learning on heterogeneous systems,
  2015.
\newblock URL \url{https://www.tensorflow.org/}.
\newblock Software available from tensorflow.org.

\bibitem[Addanki et~al.(2019)Addanki, Bojja~Venkatakrishnan, Gupta, Mao, and
  Alizadeh]{placeto}
Ravichandra Addanki, Shaileshh Bojja~Venkatakrishnan, Shreyan Gupta, Hongzi
  Mao, and Mohammad Alizadeh.
\newblock Learning generalizable device placement algorithms for distributed
  machine learning.
\newblock In H.~Wallach, H.~Larochelle, A.~Beygelzimer, F.~d`Alch\'{e} Buc,
  E.~Fox, and R.~Garnett, editors, \emph{Advances in Neural Information
  Processing Systems}, volume~32. Curran Associates, Inc., 2019.
\newblock URL
  \url{https://proceedings.neurips.cc/paper/2019/file/71560ce98c8250ce57a6a970c9991a5f-Paper.pdf}.

\bibitem[Babuschkin et~al.(2020)Babuschkin, Baumli, Bell, Bhupatiraju, Bruce,
  Buchlovsky, Budden, Cai, Clark, Danihelka, Fantacci, Godwin, Jones, Hennigan,
  Hessel, Kapturowski, Keck, Kemaev, King, Martens, Mikulik, Norman, Quan,
  Papamakarios, Ring, Ruiz, Sanchez, Schneider, Sezener, Spencer, Srinivasan,
  Stokowiec, and Viola]{deepmind2020jax}
Igor Babuschkin, Kate Baumli, Alison Bell, Surya Bhupatiraju, Jake Bruce, Peter
  Buchlovsky, David Budden, Trevor Cai, Aidan Clark, Ivo Danihelka, Claudio
  Fantacci, Jonathan Godwin, Chris Jones, Tom Hennigan, Matteo Hessel, Steven
  Kapturowski, Thomas Keck, Iurii Kemaev, Michael King, Lena Martens, Vladimir
  Mikulik, Tamara Norman, John Quan, George Papamakarios, Roman Ring, Francisco
  Ruiz, Alvaro Sanchez, Rosalia Schneider, Eren Sezener, Stephen Spencer,
  Srivatsan Srinivasan, Wojciech Stokowiec, and Fabio Viola.
\newblock The {D}eep{M}ind {JAX} {E}cosystem, 2020.
\newblock URL \url{http://github.com/deepmind}.

\bibitem[Battaglia et~al.(2016)Battaglia, Pascanu, Lai, Rezende, and
  Kavukcuoglu]{DBLP:journals/corr/BattagliaPLRK16}
Peter~W. Battaglia, Razvan Pascanu, Matthew Lai, Danilo~Jimenez Rezende, and
  Koray Kavukcuoglu.
\newblock Interaction networks for learning about objects, relations and
  physics.
\newblock \emph{CoRR}, abs/1612.00222, 2016.
\newblock URL \url{http://arxiv.org/abs/1612.00222}.

\bibitem[Bradbury et~al.(2018)Bradbury, Frostig, Hawkins, Johnson, Leary,
  Maclaurin, Necula, Paszke, Vander{P}las, Wanderman-{M}ilne, and
  Zhang]{jax2018github}
James Bradbury, Roy Frostig, Peter Hawkins, Matthew~James Johnson, Chris Leary,
  Dougal Maclaurin, George Necula, Adam Paszke, Jake Vander{P}las, Skye
  Wanderman-{M}ilne, and Qiao Zhang.
\newblock {JAX}: composable transformations of {P}ython+{N}um{P}y programs,
  2018.
\newblock URL \url{http://github.com/google/jax}.

\bibitem[Brown et~al.(2020)Brown, Mann, Ryder, Subbiah, Kaplan, Dhariwal,
  Neelakantan, Shyam, Sastry, Askell, Agarwal, Herbert{-}Voss, Krueger,
  Henighan, Child, Ramesh, Ziegler, Wu, Winter, Hesse, Chen, Sigler, Litwin,
  Gray, Chess, Clark, Berner, McCandlish, Radford, Sutskever, and
  Amodei]{gpt3_2020}
Tom~B. Brown, Benjamin Mann, Nick Ryder, Melanie Subbiah, Jared Kaplan,
  Prafulla Dhariwal, Arvind Neelakantan, Pranav Shyam, Girish Sastry, Amanda
  Askell, Sandhini Agarwal, Ariel Herbert{-}Voss, Gretchen Krueger, Tom
  Henighan, Rewon Child, Aditya Ramesh, Daniel~M. Ziegler, Jeffrey Wu, Clemens
  Winter, Christopher Hesse, Mark Chen, Eric Sigler, Mateusz Litwin, Scott
  Gray, Benjamin Chess, Jack Clark, Christopher Berner, Sam McCandlish, Alec
  Radford, Ilya Sutskever, and Dario Amodei.
\newblock Language models are few-shot learners.
\newblock \emph{CoRR}, abs/2005.14165, 2020.
\newblock URL \url{https://arxiv.org/abs/2005.14165}.

\bibitem[Browne et~al.(2012)Browne, Powley, Whitehouse, Lucas, Cowling,
  Tavener, Perez, Samothrakis, Colton, and et~al.]{Browne12asurvey}
Cameron Browne, Edward Powley, Daniel Whitehouse, Simon Lucas, Peter~I.
  Cowling, Stephen Tavener, Diego Perez, Spyridon Samothrakis, Simon Colton,
  and et~al.
\newblock A survey of monte carlo tree search methods.
\newblock \emph{IEEE TRANSACTIONS ON COMPUTATIONAL INTELLIGENCE AND AI}, 2012.

\bibitem[Chen et~al.(2021)Chen, Tworek, Jun, Yuan, de~Oliveira~Pinto, Kaplan,
  Edwards, Burda, Joseph, Brockman, Ray, Puri, Krueger, Petrov, Khlaaf, Sastry,
  Mishkin, Chan, Gray, Ryder, Pavlov, Power, Kaiser, Bavarian, Winter, Tillet,
  Such, Cummings, Plappert, Chantzis, Barnes, Herbert{-}Voss, Guss, Nichol,
  Paino, Tezak, Tang, Babuschkin, Balaji, Jain, Saunders, Hesse, Carr, Leike,
  Achiam, Misra, Morikawa, Radford, Knight, Brundage, Murati, Mayer, Welinder,
  McGrew, Amodei, McCandlish, Sutskever, and Zaremba]{codex2021}
Mark Chen, Jerry Tworek, Heewoo Jun, Qiming Yuan, Henrique~Ponde
  de~Oliveira~Pinto, Jared Kaplan, Harrison Edwards, Yuri Burda, Nicholas
  Joseph, Greg Brockman, Alex Ray, Raul Puri, Gretchen Krueger, Michael Petrov,
  Heidy Khlaaf, Girish Sastry, Pamela Mishkin, Brooke Chan, Scott Gray, Nick
  Ryder, Mikhail Pavlov, Alethea Power, Lukasz Kaiser, Mohammad Bavarian,
  Clemens Winter, Philippe Tillet, Felipe~Petroski Such, Dave Cummings,
  Matthias Plappert, Fotios Chantzis, Elizabeth Barnes, Ariel Herbert{-}Voss,
  William~Hebgen Guss, Alex Nichol, Alex Paino, Nikolas Tezak, Jie Tang, Igor
  Babuschkin, Suchir Balaji, Shantanu Jain, William Saunders, Christopher
  Hesse, Andrew~N. Carr, Jan Leike, Joshua Achiam, Vedant Misra, Evan Morikawa,
  Alec Radford, Matthew Knight, Miles Brundage, Mira Murati, Katie Mayer, Peter
  Welinder, Bob McGrew, Dario Amodei, Sam McCandlish, Ilya Sutskever, and
  Wojciech Zaremba.
\newblock Evaluating large language models trained on code.
\newblock \emph{CoRR}, abs/2107.03374, 2021.
\newblock URL \url{https://arxiv.org/abs/2107.03374}.

\bibitem[Chen et~al.(2018)Chen, Moreau, Jiang, Shen, Yan, Wang, Hu, Ceze,
  Guestrin, and Krishnamurthy]{tvm2018}
Tianqi Chen, Thierry Moreau, Ziheng Jiang, Haichen Shen, Eddie~Q. Yan, Leyuan
  Wang, Yuwei Hu, Luis Ceze, Carlos Guestrin, and Arvind Krishnamurthy.
\newblock {TVM:} end-to-end optimization stack for deep learning.
\newblock \emph{CoRR}, abs/1802.04799, 2018.
\newblock URL \url{http://arxiv.org/abs/1802.04799}.

\bibitem[Godwin* et~al.(2020)Godwin*, Keck*, Battaglia, Bapst, Kipf, Li,
  Stachenfeld, Veli\v{c}kovi\'{c}, and Sanchez-Gonzalez]{jraph2020github}
Jonathan Godwin*, Thomas Keck*, Peter Battaglia, Victor Bapst, Thomas Kipf,
  Yujia Li, Kimberly Stachenfeld, Petar Veli\v{c}kovi\'{c}, and Alvaro
  Sanchez-Gonzalez.
\newblock {J}raph: {A} library for graph neural networks in jax., 2020.
\newblock URL \url{http://github.com/deepmind/jraph}.

\bibitem[Hennigan et~al.(2020)Hennigan, Cai, Norman, and
  Babuschkin]{haiku2020github}
Tom Hennigan, Trevor Cai, Tamara Norman, and Igor Babuschkin.
\newblock {H}aiku: {S}onnet for {JAX}, 2020.
\newblock URL \url{http://github.com/deepmind/dm-haiku}.

\bibitem[Huang et~al.(2019)Huang, Cheng, Bapna, Firat, Chen, Chen, Lee, Ngiam,
  Le, Wu, and Chen]{gpipe}
Yanping Huang, Youlong Cheng, Ankur Bapna, Orhan Firat, Dehao Chen, Mia~Xu
  Chen, HyoukJoong Lee, Jiquan Ngiam, Quoc~V. Le, Yonghui Wu, and Zhifeng Chen.
\newblock Gpipe: Efficient training of giant neural networks using pipeline
  parallelism.
\newblock In Hanna~M. Wallach, Hugo Larochelle, Alina Beygelzimer, Florence
  d'Alch{\'{e}}{-}Buc, Emily~B. Fox, and Roman Garnett, editors, \emph{Advances
  in Neural Information Processing Systems 32: Annual Conference on Neural
  Information Processing Systems 2019, NeurIPS 2019, December 8-14, 2019,
  Vancouver, BC, Canada}, pages 103--112, 2019.
\newblock URL
  \url{https://proceedings.neurips.cc/paper/2019/hash/093f65e080a295f8076b1c5722a46aa2-Abstract.html}.

\bibitem[Jia et~al.(2019)Jia, Zaharia, and Aiken]{flexflow2019}
Zhihao Jia, Matei Zaharia, and Alex Aiken.
\newblock Beyond data and model parallelism for deep neural networks.
\newblock In Ameet Talwalkar, Virginia Smith, and Matei Zaharia, editors,
  \emph{Proceedings of Machine Learning and Systems 2019, MLSys 2019, Stanford,
  CA, USA, March 31 - April 2, 2019}. mlsys.org, 2019.
\newblock URL \url{https://proceedings.mlsys.org/book/265.pdf}.

\bibitem[Jia et~al.(2020)Jia, Lin, Gao, Zaharia, and Aiken]{roc2020}
Zhihao Jia, Sina Lin, Mingyu Gao, Matei Zaharia, and Alex Aiken.
\newblock Improving the accuracy, scalability, and performance of graph neural
  networks with roc.
\newblock In I.~Dhillon, D.~Papailiopoulos, and V.~Sze, editors,
  \emph{Proceedings of Machine Learning and Systems}, volume~2, pages 187--198,
  2020.
\newblock URL
  \url{https://proceedings.mlsys.org/paper/2020/file/fe9fc289c3ff0af142b6d3bead98a923-Paper.pdf}.

\bibitem[Jouppi et~al.(2017)Jouppi, Young, Patil, Patterson, Agrawal, Bajwa,
  Bates, Bhatia, Boden, Borchers, Boyle, Cantin, Chao, Clark, Coriell, Daley,
  Dau, Dean, Gelb, Ghaemmaghami, Gottipati, Gulland, Hagmann, Ho, Hogberg, Hu,
  Hundt, Hurt, Ibarz, Jaffey, Jaworski, Kaplan, Khaitan, Koch, Kumar, Lacy,
  Laudon, Law, Le, Leary, Liu, Lucke, Lundin, MacKean, Maggiore, Mahony,
  Miller, Nagarajan, Narayanaswami, Ni, Nix, Norrie, Omernick, Penukonda,
  Phelps, Ross, Salek, Samadiani, Severn, Sizikov, Snelham, Souter, Steinberg,
  Swing, Tan, Thorson, Tian, Toma, Tuttle, Vasudevan, Walter, Wang, Wilcox, and
  Yoon]{DBLP:journals/corr/JouppiYPPABBBBB17}
Norman~P. Jouppi, Cliff Young, Nishant Patil, David~A. Patterson, Gaurav
  Agrawal, Raminder Bajwa, Sarah Bates, Suresh Bhatia, Nan Boden, Al~Borchers,
  Rick Boyle, Pierre{-}luc Cantin, Clifford Chao, Chris Clark, Jeremy Coriell,
  Mike Daley, Matt Dau, Jeffrey Dean, Ben Gelb, Tara~Vazir Ghaemmaghami,
  Rajendra Gottipati, William Gulland, Robert Hagmann, Richard~C. Ho, Doug
  Hogberg, John Hu, Robert Hundt, Dan Hurt, Julian Ibarz, Aaron Jaffey, Alek
  Jaworski, Alexander Kaplan, Harshit Khaitan, Andy Koch, Naveen Kumar, Steve
  Lacy, James Laudon, James Law, Diemthu Le, Chris Leary, Zhuyuan Liu, Kyle
  Lucke, Alan Lundin, Gordon MacKean, Adriana Maggiore, Maire Mahony, Kieran
  Miller, Rahul Nagarajan, Ravi Narayanaswami, Ray Ni, Kathy Nix, Thomas
  Norrie, Mark Omernick, Narayana Penukonda, Andy Phelps, Jonathan Ross, Amir
  Salek, Emad Samadiani, Chris Severn, Gregory Sizikov, Matthew Snelham, Jed
  Souter, Dan Steinberg, Andy Swing, Mercedes Tan, Gregory Thorson, Bo~Tian,
  Horia Toma, Erick Tuttle, Vijay Vasudevan, Richard Walter, Walter Wang, Eric
  Wilcox, and Doe~Hyun Yoon.
\newblock In-datacenter performance analysis of a tensor processing unit.
\newblock \emph{CoRR}, abs/1704.04760, 2017.
\newblock URL \url{http://arxiv.org/abs/1704.04760}.

\bibitem[Kaplan et~al.(2020)Kaplan, McCandlish, Henighan, Brown, Chess, Child,
  Gray, Radford, Wu, and Amodei]{scaling_laws_2020}
Jared Kaplan, Sam McCandlish, Tom Henighan, Tom~B. Brown, Benjamin Chess, Rewon
  Child, Scott Gray, Alec Radford, Jeffrey Wu, and Dario Amodei.
\newblock Scaling laws for neural language models.
\newblock \emph{CoRR}, abs/2001.08361, 2020.
\newblock URL \url{https://arxiv.org/abs/2001.08361}.

\bibitem[Lattner et~al.(2021)Lattner, Amini, Bondhugula, Cohen, Davis, Pienaar,
  Riddle, Shpeisman, Vasilache, and Zinenko]{mlir2020}
Chris Lattner, Mehdi Amini, Uday Bondhugula, Albert Cohen, Andy Davis, Jacques
  Pienaar, River Riddle, Tatiana Shpeisman, Nicolas Vasilache, and Oleksandr
  Zinenko.
\newblock Mlir: Scaling compiler infrastructure for domain specific
  computation.
\newblock In \emph{2021 IEEE/ACM International Symposium on Code Generation and
  Optimization (CGO)}, pages 2--14, 2021.
\newblock \doi{10.1109/CGO51591.2021.9370308}.

\bibitem[Lepikhin et~al.(2020)Lepikhin, Lee, Xu, Chen, Firat, Huang, Krikun,
  Shazeer, and Chen]{gshard}
Dmitry Lepikhin, HyoukJoong Lee, Yuanzhong Xu, Dehao Chen, Orhan Firat, Yanping
  Huang, Maxim Krikun, Noam Shazeer, and Zhifeng Chen.
\newblock Gshard: Scaling giant models with conditional computation and
  automatic sharding.
\newblock \emph{CoRR}, abs/2006.16668, 2020.
\newblock URL \url{https://arxiv.org/abs/2006.16668}.

\bibitem[Mirhoseini et~al.(2017)Mirhoseini, Pham, Le, Norouzi, Bengio, Steiner,
  Zhou, Kumar, Larsen, and Dean]{deviceplacement2017}
Azalia Mirhoseini, Hieu Pham, Quoc Le, Mohammad Norouzi, Samy Bengio, Benoit
  Steiner, Yuefeng Zhou, Naveen Kumar, Rasmus Larsen, and Jeff Dean.
\newblock Device placement optimization with reinforcement learning, 2017.
\newblock URL \url{https://arxiv.org/abs/1706.04972}.

\bibitem[Mirhoseini et~al.(2018)Mirhoseini, Goldie, Pham, Steiner, Le, and
  Dean]{mirhoseini2018a}
Azalia Mirhoseini, Anna Goldie, Hieu Pham, Benoit Steiner, Quoc~V. Le, and Jeff
  Dean.
\newblock A hierarchical model for device placement.
\newblock In \emph{International Conference on Learning Representations}, 2018.
\newblock URL \url{https://openreview.net/forum?id=Hkc-TeZ0W}.

\bibitem[Narayanan et~al.(2019)Narayanan, Harlap, Phanishayee, Seshadri,
  Devanur, Ganger, Gibbons, and Zaharia]{pipedream}
Deepak Narayanan, Aaron Harlap, Amar Phanishayee, Vivek Seshadri, Nikhil~R.
  Devanur, Gregory~R. Ganger, Phillip~B. Gibbons, and Matei Zaharia.
\newblock Pipedream: generalized pipeline parallelism for {DNN} training.
\newblock In Tim Brecht and Carey Williamson, editors, \emph{Proceedings of the
  27th {ACM} Symposium on Operating Systems Principles, {SOSP} 2019,
  Huntsville, ON, Canada, October 27-30, 2019}, pages 1--15. {ACM}, 2019.
\newblock \doi{10.1145/3341301.3359646}.
\newblock URL \url{https://doi.org/10.1145/3341301.3359646}.

\bibitem[Narayanan et~al.(2021{\natexlab{a}})Narayanan, Phanishayee, Shi, Chen,
  and Zaharia]{Narayanan2021MemoryEfficientPD}
Deepak Narayanan, Amar Phanishayee, Kaiyu Shi, Xie Chen, and Matei~A. Zaharia.
\newblock Memory-efficient pipeline-parallel dnn training.
\newblock In \emph{ICML}, 2021{\natexlab{a}}.

\bibitem[Narayanan et~al.(2021{\natexlab{b}})Narayanan, Shoeybi, Casper,
  LeGresley, Patwary, Korthikanti, Vainbrand, Kashinkunti, Bernauer, Catanzaro,
  Phanishayee, and Zaharia]{large_megatron_21}
Deepak Narayanan, Mohammad Shoeybi, Jared Casper, Patrick LeGresley, Mostofa
  Patwary, Vijay Korthikanti, Dmitri Vainbrand, Prethvi Kashinkunti, Julie
  Bernauer, Bryan Catanzaro, Amar Phanishayee, and Matei Zaharia.
\newblock Efficient large-scale language model training on {GPU} clusters.
\newblock \emph{CoRR}, abs/2104.04473, 2021{\natexlab{b}}.
\newblock URL \url{https://arxiv.org/abs/2104.04473}.

\bibitem[Paliwal et~al.(2020)Paliwal, Gimeno, Nair, Li, Lubin, Kohli, and
  Vinyals]{aditya2020}
Aditya Paliwal, Felix Gimeno, Vinod Nair, Yujia Li, Miles Lubin, Pushmeet
  Kohli, and Oriol Vinyals.
\newblock Reinforced genetic algorithm learning for optimizing computation
  graphs.
\newblock In \emph{ICLR}, 2020.

\bibitem[Paszke et~al.(2019)Paszke, Gross, Massa, Lerer, Bradbury, Chanan,
  Killeen, Lin, Gimelshein, Antiga, Desmaison, Kopf, Yang, DeVito, Raison,
  Tejani, Chilamkurthy, Steiner, Fang, Bai, and Chintala]{pytorch2019}
Adam Paszke, Sam Gross, Francisco Massa, Adam Lerer, James Bradbury, Gregory
  Chanan, Trevor Killeen, Zeming Lin, Natalia Gimelshein, Luca Antiga, Alban
  Desmaison, Andreas Kopf, Edward Yang, Zachary DeVito, Martin Raison, Alykhan
  Tejani, Sasank Chilamkurthy, Benoit Steiner, Lu~Fang, Junjie Bai, and Soumith
  Chintala.
\newblock Pytorch: An imperative style, high-performance deep learning library.
\newblock In H.~Wallach, H.~Larochelle, A.~Beygelzimer, F.~d`Alch\'{e} Buc,
  E.~Fox, and R.~Garnett, editors, \emph{Advances in Neural Information
  Processing Systems}, volume~32. Curran Associates, Inc., 2019.
\newblock URL
  \url{https://proceedings.neurips.cc/paper/2019/file/bdbca288fee7f92f2bfa9f7012727740-Paper.pdf}.

\bibitem[Ren et~al.(2021)Ren, Rajbhandari, Aminabadi, Ruwase, Yang, Zhang, Li,
  and He]{DBLP:journals/corr/abs-2101-06840}
Jie Ren, Samyam Rajbhandari, Reza~Yazdani Aminabadi, Olatunji Ruwase, Shuangyan
  Yang, Minjia Zhang, Dong Li, and Yuxiong He.
\newblock Zero-offload: Democratizing billion-scale model training.
\newblock \emph{CoRR}, abs/2101.06840, 2021.
\newblock URL \url{https://arxiv.org/abs/2101.06840}.

\bibitem[Santhanam et~al.(2021)Santhanam, Krishna, Tomioka, Fitzgibbon, and
  Harris]{distir2021}
Keshav Santhanam, Siddharth Krishna, Ryota Tomioka, Andrew Fitzgibbon, and Tim
  Harris.
\newblock Distir: An intermediate representation for optimizing distributed
  neural networks.
\newblock In \emph{Proceedings of the 1st Workshop on Machine Learning and
  Systems}, EuroMLSys '21, page 15–23, New York, NY, USA, 2021. Association
  for Computing Machinery.
\newblock ISBN 9781450382984.
\newblock \doi{10.1145/3437984.3458829}.
\newblock URL \url{https://doi.org/10.1145/3437984.3458829}.

\bibitem[Shoeybi et~al.(2019)Shoeybi, Patwary, Puri, LeGresley, Casper, and
  Catanzaro]{megatron2019}
Mohammad Shoeybi, Mostofa Patwary, Raul Puri, Patrick LeGresley, Jared Casper,
  and Bryan Catanzaro.
\newblock Megatron-lm: Training multi-billion parameter language models using
  model parallelism.
\newblock \emph{CoRR}, abs/1909.08053, 2019.
\newblock URL \url{http://arxiv.org/abs/1909.08053}.

\bibitem[Vaswani et~al.(2017)Vaswani, Shazeer, Parmar, Uszkoreit, Jones, Gomez,
  Kaiser, and Polosukhin]{DBLP:journals/corr/VaswaniSPUJGKP17}
Ashish Vaswani, Noam Shazeer, Niki Parmar, Jakob Uszkoreit, Llion Jones,
  Aidan~N. Gomez, Lukasz Kaiser, and Illia Polosukhin.
\newblock Attention is all you need.
\newblock \emph{CoRR}, abs/1706.03762, 2017.
\newblock URL \url{http://arxiv.org/abs/1706.03762}.

\bibitem[Vytiniotis(2020)]{partir2020}
Dimitrios Vytiniotis.
\newblock Declarative abstractions for tensor program partitioning.
\newblock In \emph{Proceedings of the 22nd International Symposium on
  Principles and Practice of Declarative Programming}, PPDP '20, New York, NY,
  USA, 2020. Association for Computing Machinery.
\newblock ISBN 9781450388214.
\newblock \doi{10.1145/3414080.3414105}.
\newblock URL \url{https://doi.org/10.1145/3414080.3414105}.

\bibitem[Wang et~al.(2019)Wang, Huang, and Li]{tofu2019}
Minjie Wang, Chien-chin Huang, and Jinyang Li.
\newblock Supporting very large models using automatic dataflow graph
  partitioning.
\newblock In \emph{Proceedings of the Fourteenth EuroSys Conference 2019},
  EuroSys '19, New York, NY, USA, 2019. Association for Computing Machinery.
\newblock ISBN 9781450362818.
\newblock \doi{10.1145/3302424.3303953}.
\newblock URL \url{https://doi.org/10.1145/3302424.3303953}.

\bibitem[Xu et~al.(2021)Xu, Lee, Chen, Hechtman, Huang, Joshi, Krikun,
  Lepikhin, Ly, Maggioni, Pang, Shazeer, Wang, Wang, Wu, and Chen]{gspmd2021}
Yuanzhong Xu, HyoukJoong Lee, Dehao Chen, Blake~A. Hechtman, Yanping Huang,
  Rahul Joshi, Maxim Krikun, Dmitry Lepikhin, Andy Ly, Marcello Maggioni,
  Ruoming Pang, Noam Shazeer, Shibo Wang, Tao Wang, Yonghui Wu, and Zhifeng
  Chen.
\newblock {GSPMD:} general and scalable parallelization for {ML} computation
  graphs.
\newblock \emph{CoRR}, abs/2105.04663, 2021.
\newblock URL \url{https://arxiv.org/abs/2105.04663}.

\bibitem[Yang et~al.(2019)Yang, Zhang, Li, R{\'{e}}, Aberger, and
  Sa]{pipemare21}
Bowen Yang, Jian Zhang, Jonathan Li, Christopher R{\'{e}}, Christopher~R.
  Aberger, and Christopher~De Sa.
\newblock Pipemare: Asynchronous pipeline parallel {DNN} training.
\newblock \emph{CoRR}, abs/1910.05124, 2019.
\newblock URL \url{http://arxiv.org/abs/1910.05124}.

\bibitem[Yang et~al.(2021)Yang, Phothilimthana, Wang, Willsey, Roy, and
  Pienaar]{equality_saturation21}
Yichen Yang, Phitchaya Phothilimthana, Yisu Wang, Max Willsey, Sudip Roy, and
  Jacques Pienaar.
\newblock Equality saturation for tensor graph superoptimization.
\newblock In A.~Smola, A.~Dimakis, and I.~Stoica, editors, \emph{Proceedings of
  Machine Learning and Systems}, volume~3, pages 255--268, 2021.
\newblock URL
  \url{https://proceedings.mlsys.org/paper/2021/file/65ded5353c5ee48d0b7d48c591b8f430-Paper.pdf}.

\bibitem[Zhou et~al.(2020)Zhou, Roy, Abdolrashidi, Wong, Ma, Xu, Liu,
  Phothilimtha, Wang, Goldie, Mirhoseini, and Laudon]{optgraphcomp}
Yanqi Zhou, Sudip Roy, Amirali Abdolrashidi, Daniel Wong, Peter Ma, Qiumin Xu,
  Hanxiao Liu, Phitchaya Phothilimtha, Shen Wang, Anna Goldie, Azalia
  Mirhoseini, and James Laudon.
\newblock Transferable graph optimizers for ml compilers.
\newblock In H.~Larochelle, M.~Ranzato, R.~Hadsell, M.~F. Balcan, and H.~Lin,
  editors, \emph{Advances in Neural Information Processing Systems}, volume~33,
  pages 13844--13855. Curran Associates, Inc., 2020.
\newblock URL
  \url{https://proceedings.neurips.cc/paper/2020/file/9f29450d2eb58feb555078bdefe28aa5-Paper.pdf}.

\end{thebibliography}
\appendix
\end{document}